\documentclass{svproc}
\usepackage{graphicx}            %
\usepackage{color}
\usepackage{amsmath}             %
\usepackage{braket}
\usepackage{amssymb}             %
\usepackage{amsfonts}            %
\usepackage{enumitem}            %
\usepackage{multirow}            %
\usepackage{siunitx}             %
\usepackage{url}                 %

\usepackage{xspace}              %
\usepackage[T1]{fontenc}         %
\usepackage[hidelinks]{hyperref} %
\usepackage{balance}
\usepackage[bottom]{footmisc}
\usepackage{xfrac}
\usepackage{tensor}
\usepackage{todonotes}

\usepackage[firstpage=true]{background}
\newcommand\copyrighttext{%
\parbox{1.55\textwidth}{
\footnotesize
25th International Conference Series on Climbing and Walking Robots (CLAWAR), Ponta Delgada, Azores, Portugal, September 2022.
}}

\SetBgContents{\copyrighttext}
\SetBgScale{1}
\SetBgColor{black}
\SetBgAngle{0}
\SetBgOpacity{1}
\SetBgPosition{current page.north}
\SetBgVshift{-0.8cm}

\usepackage[bottom]{footmisc}

\graphicspath{{images/}}
\DeclareGraphicsExtensions{.pdf,.png,.jpg,.jpeg}

\DeclareMathOperator{\sgn}{sgn}

\newcommand{\mypm}{\mathbin{\mathpalette\@mypm\relax}}

\makeatletter
\newcommand{\amsray}{%
\mathpalette {\overarrow@\rayfill@}}
\def\rayfill@{\arrowfill@{\mkern4mu\mapstochar\relbar}\relbar{\mkern 4.08mu}}%
\makeatother

\newcommand{\seclabel}[1]{\label{sec:#1}}

\newcommand{\figlabel}[1]{\label{fig:#1}}
\newcommand{\tablabel}[1]{\label{tab:#1}}
\newcommand{\eqnlabel}[1]{\label{eqn:#1}}

\newcommand{\secref}[1]{Section~\ref{sec:#1}\xspace}

\newcommand{\figref}[1]{Fig.~\ref{fig:#1}\xspace}

\newcommand{\tabref}[1]{Table~\ref{tab:#1}\xspace}
\newcommand{\eqnref}[1]{(\ref{eqn:#1})\xspace}

\newcommand{\iguhop}{igus\textsuperscript{\tiny\circledR}$\!$ Humanoid Open Platform\xspace}

\begin{document}
\mainmatter
\title{Direct Centroidal Control\\for Balanced Humanoid Locomotion}
\titlerunning{Direct Centroidal Control for Balanced Humanoid Locomotion}
\author{Grzegorz Ficht \and Sven Behnke}
\authorrunning{Ficht, Behnke} %
\tocauthor{Grzegorz Ficht and Sven Behnke}
\institute{Institute for Computer Science VI, University of Bonn\\
Endenicher Allee 19a, 53115 Bonn, Germany\\
\email{\{ficht, behnke\}@ais.uni-bonn.de}}
\maketitle
\begin{abstract}
We present an integrated approach to locomotion and balancing of humanoid robots based on direct centroidal control.
Our method uses a five-mass description of a humanoid. It generates whole-body motions from desired foot trajectories and centroidal 
parameters of the robot. A set of simplified models is used to formulate general and intuitive control laws, which are then 
applied in real-time for estimating and regulating the center of mass position and orientation of the multibody's principal 
axes of inertia. The combination of proposed algorithms produces a stretched-leg gait with naturally looking upper body motions.
As only a 6-axis IMU and joint encoders are necessary for the implementation, the portability between robots is high.
Our method has been experimentally verified using an \iguhop, demonstrating whole-body locomotion and push rejection
capabilities. 
\end{abstract}

\section{Introduction}
Dynamic control of humanoid robots has been an active research area for several years. The high-dimensionality and underactuation 
of the system, coupled with environment unpredictability, signal noise, and non-deterministic actuation has led to the development of numerous 
approaches. When real-time control poses no issue, task-based optimal control methods are capable of producing 
motion plans exploiting the full dynamics with constraint awareness~\cite{al2012trajectory,lengagne2013generation}.

Supporting online application has been possible by utilising models capable of generalising a core part of the dynamics.
For humanoid walking and balancing, a substantial line of work is based on tracking Center of Mass~(CoM) and Zero Moment Point~(ZMP) trajectories, 
generated with the Linear Inverted Pendulum Model~(LIPM). Given desired ZMP locations, a CoM trajectory
can be generated using preview control~\cite{kajita2003biped} and tracked in a state-space setting~\cite{kajita2010biped}. 
Computing optimal control gains---or even the complete trajectories over long horizons---can be done in a 
Linear-Quadratic Regulator~(LQR) setting~\cite{tedrake2015closed}. Another approach involves using Model Predictive 
Control~(MPC)~\cite{wieber2006trajectory,dimitrov2011sparse}, also capable of handling large perturbations. 
Alternatively, some works build the control around the Capture Point~(CP), reducing the system dynamics to first order~\cite{englsberger2011bipedal,englsberger2015three}.

\begin{figure}[!t]
\centering{\includegraphics[width=0.88\linewidth]{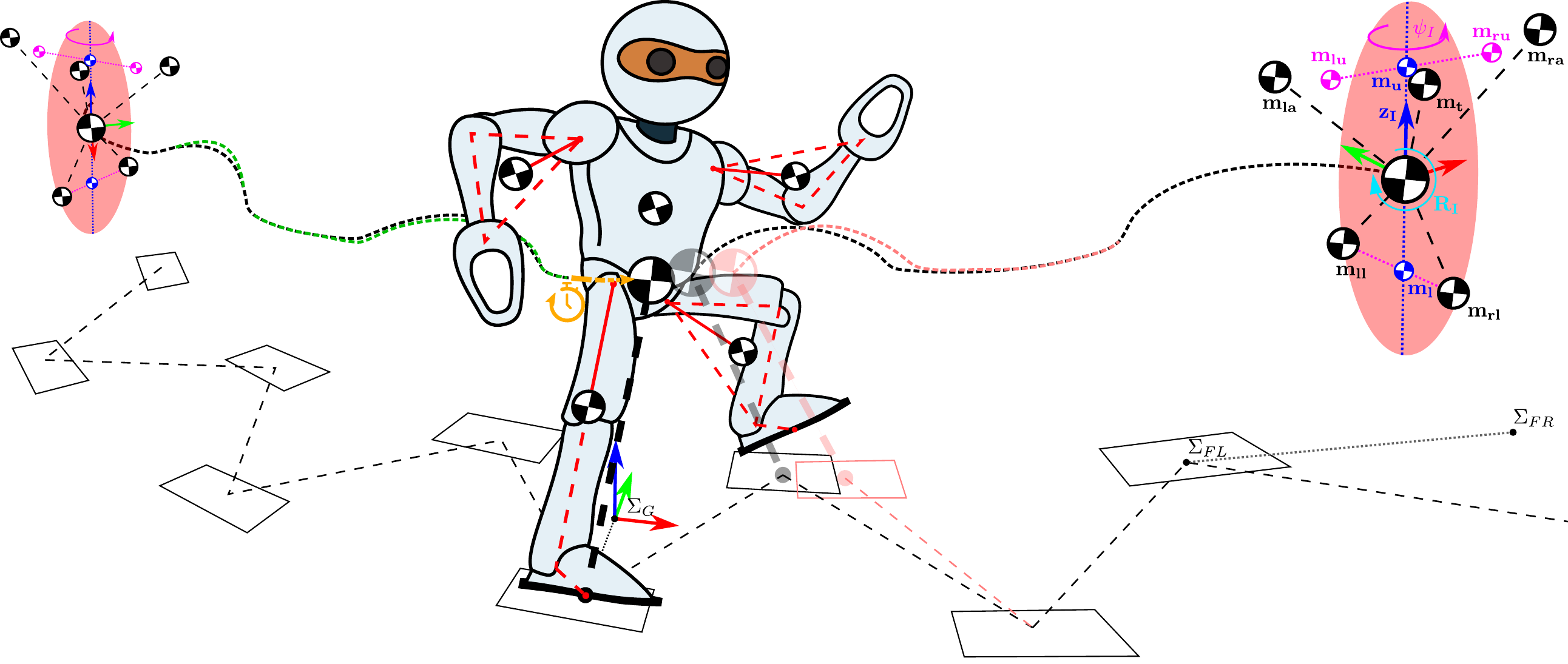}}
\vspace{-3ex}
\caption{Direct centroidal control, including CoM and CAM regulation with step size and timing adjustment. Our approach 
achieves walking with straight legs and angular momentum regulation using the tilt and yaw of the inertia principal axes. Five masses 
are combined into three dumbbels, whose relative movement translates between the full humanoid kinematics and its centroidal model.
}
\figlabel{teaser} 
\vspace{-4ex}
\end{figure}

The LIPM, while extremely useful, has two limitations which can hinder walking performance. 
The first one is constraining the CoM to move on a horizontal plane, which requires it's controllability in the vertical axis. Ensuring this
usually results in the well-known, bent-knee walking to avoid kinematic singularities. 
Improving this issue has been the aim of multiple research works~\cite{li2010trajectory,griffin2018straight,kajita2019position}. 
The second limitation stems from the assumption of zero angular momentum around the CoM. This is not an issue for
generating the CoM motion, as humans are known to regulate the Centroidal Angular Momentum~(CAM) close to zero when 
walking~\cite{popovic2004angular}. This ability however, needs to be included in the control
and has led to the flywheel-extended LIPM~\cite{pratt2006capture}. The legs
are then often assumed to be massless, while the torso acts as the flywheel~\cite{whitman2012torso}. This approximation is favored as
so far there has been no clear and intuitive angular counterpart to the linear CoM motion. To the best 
of our knowledge, only angular excursions from whole body angular velocities so far have been used~\cite{popovic2004angular}. %
Simultaneous control of linear and angular momenta has been tackled also in more general approaches. Examples include the Resolved 
Momentum Control framework~\cite{kajita2003resolved} and Whole-Body Control~(WBC) for motion generation using centroidal dynamics~\cite{orin2013centroidal}, 
with improving online application being an active research topic~\cite{dai2014whole,wensing2016improved}.

A separate consideration to the control method is the capability of the hardware to execute these motion plans. On the hardware layer, this is 
often realized with backlash-free reducers and a high gain setting, to assure joint tracking performance. In lower-quality 
hardware, the limited joint torque and presence of compliance, backlash, and latency render many of these methods unapplicable.
Also, ground contact and stability assessment is often assumed as given with force and/or torque sensors, which is not always the case~\cite{masuya2020review}.
In this regard, there exists a whole ZMP-less category of methods, which are designed around biologically-inspired 
Central Pattern Generators~(CPG)~\cite{nielsen2003we}. As the trajectories are manually designed to produce a \textit{self-stable}, open-loop gait,
no precise modelling or foot sensors are required. Push rejection can be implemented
on the joint level~\cite{allgeuer2018bipedal} or by modifying gait commands to alter the step parameters~\cite{missura2019capture}.

The contribution of this paper is an approach to whole-body humanoid robot locomotion by generating motion frames approximating the 
Composite Rigid Body~(CRB) Inertia with five masses. To improve tracking performance, 
knee torque is limited through the inclusion of a straight-support-leg constraint. Reference CoM and ZMP values are combined with 
foot trajectories to determine reference orientations of the principle axes of inertia $\mathbf{R}_\mathbf{I}$, which we introduce as 
an angular position equivalent to enhance the walking capabilities through CAM regulation. Control laws are then extended to track 
the centroidal parameters for both the linear and angular movement. Additionally, step size and timing feedback is developed using a 
closed-form prediction of the end-of-step state. As the approach requires only a 6-axis IMU and joint position feedback, it can be 
applied to a wide variety of robots.

\section{Five-Mass Centroidal Description of a Humanoid} \seclabel{reducedmodel} %

As shown in~\figref{teaser}, our description of a humanoid robot is based on a non-uniform five-mass distribution model, where each limb and the torso
are represented by their respective point mass. The torso mass is offset from the center of the pelvis, assumed to be the floating 
base of the robot $\Sigma_B$. Each limb mass is attached to the torso and parameterised through a triangle approximation, which 
ties in the kinematics and mass placement with a one-to-one mapping. The relative movement of these five masses shapes the system 
centroidal properties, e.g. it's CoM and system inertia $\mathbf{I_R}$, which we used to develop a whole-body motion generator in~\cite{ficht2020fast}.
Our method can be thought of as a forward and inverse kinematics approach to whole-body humanoid control, where both the CoM and 
system inertia can be altered to achieve desired dynamic effects, e.g. concurrent linear and angular momentum regulation. 

We compute the joint configuration $\mathbf{q}$ by providing CoM-relative foot frames $\Sigma_{FL},\Sigma_{FR}$, constrained by 
set inertia principal moments $\mathbf{I_{PA}}$ and axes orientation $\mathbf{R_I}$~(\figref{teaser}, on the right). The upper 
body and lower body~(leg) masses $\mathbf{m_u}$, $\mathbf{m_l}$ form a dumbbell, which tilts around the CoM. A virtual single 
leg evaluates the placement of the base frame from $\mathbf{m_l}$ with respect to $\mathbf{m_u}$ reachability. Having confirmed a viable
dumbbell~(inertia) tilt~$\mathbf{z_I}$, the trunk is placed with its mass pointing towards $\mathbf{m_u}$ and aligned with the set inertia yaw angle $\psi_I$.
This completes the trunk orientation $\mathbf{R_t}$ for $\Sigma_B$, which in conjunction with $\Sigma_{FL}$ and $\Sigma_{FR}$ specifies 
the leg configuration. The leg masses form a dumbbell, which rotates around $\mathbf{m_l}$ with a yaw angle $\psi_l$. 
A similar dumbbell is put at $\mathbf{m_u}$, yawing with $\psi_u$. Both of these dumbbells work to realise the set yaw rotation $\psi_I$.
This top-down approach introduces more detail on each step of the pose generation scheme, which naturally resolves the upper body movement
to satisfy the constraints driven by the feet. For a full description on the pose generation scheme, please refer to~\cite{ficht2020fast}.

During the typical bent-knee walking, the majority of the weight is put on the supporting knee joint. Most tracking errors arrise at this point
due to the accumulation of non-negligible joint compliance, backlash, and limited torque. We work around this issue by extending our pose generator~\cite{ficht2020fast}
by including a straight-support-leg constraint. After computing the initial torso orientation and hip origins, we evaluate 
the distance to the supporting leg's ankle. If it is shorter than the sum of both leg links, the base frame position is shifted from the virtual 
leg origin by that difference. By doing so, the motion generator will resolve the CoM and inertia placement with a recomputed torso orientation 
$\mathbf{R_t}'$. The pose generation continues with $\Sigma_B'$, which allows to maintain control over the CoM in the vertical axis by adjusting the upper body and arms.

\section{Control Approach}  \seclabel{wbcontrol}

\figref{controlScheme} gives an overview of our approach. For brevity, we limit the majority of the equation derivations to the planar case,
three-dimensional coordinates are provided only when necessary.

\subsection{Reference Trajectory Generation} \seclabel{reference}

\begin{figure}[!t]
\centering{\includegraphics[width=0.99\linewidth]{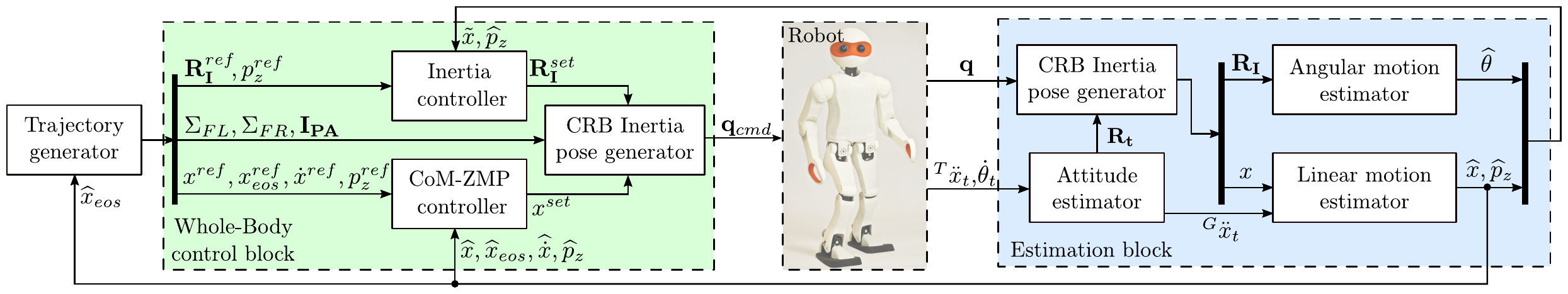}}
\vspace{-3ex}
\caption{Interaction between the framework modules.}
\figlabel{controlScheme}
\vspace{-3ex}
\end{figure}

Generating a frame of motion with the five-mass model from \secref{reducedmodel} requires specifying desired foot frames 
$\Sigma_{FL},\Sigma_{FR}$ relative to the CoM and desired system inertia $\mathbf{I_R}$ composed of the principal axes 
orientation $\mathbf{R_I}$ and moments $\mathbf{I_{PA}}$. In scenarios with plain single or double support 
balancing, the reference CoM and ZMP stay constant---unlike in walking---which requires specifying trajectories due to
contact point exchange. In this regard, we follow the approach of Kajita~\cite{kajita2019linear} with the well-known planar 
Linear Inverted Pendulum Model~(LIPM) dynamics and its analytical solution:
\begin{equation} \eqnlabel{lipm}
\begin{gathered} 
\ddot{x} = \omega^2(x-p_z), \quad \omega = \sqrt{g/h}\\
x(t) = x_0\cosh(\omega t)+\dot{x}_0\frac{1}{\omega}\sinh(\omega t),\\ 
\dot{x}(t) = x_0\omega\sinh(\omega t)+\dot{x}_0\cosh(\omega t),\\
\end{gathered} 
\end{equation}
where $h$ is the constant CoM height, $g$ is gravity, while $x$ and $p_z$ are the horizontal CoM and ZMP positions, respectively.
Given a set of initial conditions $(x_0,\dot{x}_0)$, we can compute the state $(x,\dot{x})$ with respect to elapsed time $t$. Two
independently generated planar solutions are combined into a three-dimensional ground-relative CoM coordinate $\mathbf{c}$. 
A given velocity $v^{ref}_{g}$ can be achieved through multiple combinations of step size
and applied gait frequency $f_g$. By setting a nominal frequency, we control the walking speed with stride length. The step progression is 
expressed by $\mu$: 
\begin{equation} \eqnlabel{gaitprogression}
\mu\in[0,1),\qquad \mu[n+1] = \mu[n] + f_g \Delta t,
\end{equation}
where $\Delta t$ is the time step arising from the system control frequency. We let $\mu$ wrap around when crossing 
its border values, which triggers a sign flip of the supporting leg sign $\iota:\{\text{L}=-1,\text{R}=1\}$. We compute the next initial step $\mathbf{s}_i$ coordinates 
$x_s$, $y_s$ and heading $\psi_s$ as follows:
\begin{equation}\eqnlabel{stepsize} 
\mathbf{s}_i
\begin{cases}
    x_s &= v^{ref}_{g,x}/f_g\\
    y_s &= \iota s_w + s_y=\\
    \psi_s &= v^{ref}_{g,\psi}/f_g\\
\end{cases}
\begin{cases}
2v^{ref}_{g,y}/f_g&\text{if}~\iota=\sgn(v^{ref}_{g,y}) \\
0& \text{otherwise}\\
 \end{cases},
\end{equation}
which is then refined into the applied footstep~$\mathbf{s}$ by slight position adjustments 
to produce a single continuous trajectory according to~\cite{kajita2019linear}. When moving sideways, we 
first take a larger step, and then a trailing one with nominal step width $s_w$. Only three steps are necessary to be stored at any given time. 
The supporting foot stays at the current foothold $\mathbf{s}[k]$ while the swing 
leg travels from the previous footstep location $\mathbf{s}[k-1]$ to the next one $\mathbf{s}[k+1]$, reaching a step height of $s_h$ with:
\begin{equation}\eqnlabel{foottraj} 
\Sigma_{F}
\begin{cases}
 \mathbf{f} =
\begin{cases}
    x_f = (1-\mu)\mathbf{s}^x[k-1]+\mu \mathbf{s}^x[k+1]\\
    y_f = (1-\mu)\mathbf{s}^y[k-1]+\mu \mathbf{s}^y[k+1]\\
    z_f = s_h\sin(\mu*\pi)\\
\end{cases}\\
 \psi_f = (1-\mu)\mathbf{s}^\psi[k-1]+\mu \mathbf{s}^\psi[k+1].
\end{cases}
\end{equation}
For simplicity, the roll $\phi_f$ and pitch $\theta_f$ are set to zero to keep the feet parallel to the ground.
When a foothold has been reached, the queue of steps with their coordinates are shifted, and a new step is generated. 
As the footholds are in reference to the ground, the final foot frame positions are offset by the set CoM position, 
to conform with the whole body pose generator.
The final components required to define a whole body pose are the inertial properties $\mathbf{I_R}$. The definition
for the inertia orientation $\mathbf{R}_\mathbf{I}^{ref}$ is obtained with Rodrigues' axis-angle formula from 
the inertia z-axis tilt $\mathbf{z}^{ref}_\mathbf{I}$ and the rotation angle around it $\psi_I$. A continuous \emph{neutral} 
tilt trajectory of $\mathbf{z}^{ref}_\mathbf{I}$ should not result in generating CAM. This is achieved when the torque of a 
flywheel attached to a pendulum is zero, e.g. it does not rotate with respect to it. We therefore align $\mathbf{z}^{ref}_\mathbf{I}$
with the robot tilt defined as the vector originating 
at the feet of the robot, pointing towards the CoM and set $\psi_I$ to align with the step progression $\mu$:%
\begin{equation} \eqnlabel{inertiaref}
 \mathbf{z}^{ref}_\mathbf{I} = -(\mathbf{f_l}+\mathbf{f_r})/2, \quad \psi_I = (1-\mu)\mathbf{s}^\psi[k]+\mu \mathbf{s}^\psi[k+1].
\end{equation}
With this we keep CAM mostly independent from CoM and foot placement. Finally, nominal principal moments $\mathbf{I}^{ref}_{\mathbf{PA}}$ 
are kept constant and computed from an upright standing pose, with the pose generator adjusting their values to satisfy other constraints.

\subsection{State Estimation} \seclabel{estimation} 

Estimating the robot's state is a necessity when it comes to precise control involving system dynamics. To maintain
generality and a wide applicability, we require only joint encoders and a 6-axis IMU in the trunk to reconstruct the necessary state variables. 
The IMU is used to provide the torso~(base) orientation $\mathbf{R_t}$ of $\Sigma_B$ with an attitude estimator of choice~\cite{Allgeuer2014,ludwig2018comparison,cisneros2020lyapunov}.
In combination with the measured joint angles $\mathbf{q}$, we compute the foot frames~$\Sigma_{FL},\Sigma_{FR}$ and the positions of the five masses.
These are then used to reconstruct the system CoM and inertial properties $\mathbf{I_R}$, $\mathbf{I_{PA}}$.
For synchronising the measured and control state we use a common reference \textit{ground} frame $\Sigma_G$. 
This frame is laterally offset from the supporting foot by half of the step width from the last step taken~(see \figref{teaser}).
The supporting foot is chosen as the lower one of the two with a hysteresis applied to disable rapid switching of the frames, within which the 
robot is assumed to be in \textit{semi double support}. By keeping track of every support exchange, we update the robot's odometry.
To provide the robot with a notion of balance, a stability criterion is required. As we base our control around the LIPM concept, 
we use the ZMP which is a linear combination of the CoM position $x$ and acceleration $\ddot{x}$ as seen in~\eqnref{lipm}. 
We use a Kalman filter to estimate the complete state vector $\mathbf{x} = \begin{bmatrix}x &\dot{x} &\ddot{x}\end{bmatrix}$. Unlike work
mentioned in ~\cite{masuya2020review}, we do not explicitly use the LIPM state transition matrix in the filter design to not enforce
it's dynamics in the estimation. Instead we use the linear mass model. Although the 
five-mass approximation does

provide accurate $x$ measurements, relying solely on them does not provide satisfactory acceleration estimates. ZMP estimates $\widehat{p}_z$
will be inoperative due to the non-negligible time delay before $\widehat{\ddot{x}}$ converges in response to actual changes in $\ddot{x}$. 
We supplement the measurement model $\mathbf{z}_k$ with the unrotated and unbiased for gravity $g$ trunk acceleration~$^G\ddot{x}_t$:
\begin{equation} \eqnlabel{kalmanmeasurement}
\begin{bmatrix}^G\ddot{x}_t\\ ^G\ddot{y}_t\\ ^G\ddot{z}_t\end{bmatrix} = \begin{bmatrix}^T\ddot{x}_t\\ ^T\ddot{y}_t\\ ^T\ddot{z}_t\end{bmatrix}\mathbf{R_t}-\begin{bmatrix}0\\ 0\\ g\end{bmatrix},~
\mathbf{z}_k = \begin{bmatrix}1\quad0\\0\quad0\\0\quad1\end{bmatrix}\begin{bmatrix}x\\ ^G\ddot{x}_t \end{bmatrix} + \mathbf{v}_k
\end{equation}
Having sources stemming from two different state variable measurements allows for low-delay, robust estimation of $\mathbf{x}$.
Even though $^G\ddot{x}_t$ only roughly coincides with $\ddot{x}_t$ only when CAM is zero, its fusion with CoM measurements
provides a refined estimate $\widehat{\mathbf{x}}$ with fast $\ddot{x}$ convergence. The ZMP estimate $\widehat{p}_z$ is then simply:
\begin{equation} \eqnlabel{zmpestimate}
\widehat{p}_z = \widehat{x} - \frac{\widehat{z}}{g}~\widehat{\ddot{x}}.
\end{equation}
A second Kalman filter is used for the angular motion, where angles~$\theta$ of the inertia orientation~$\mathbf{R_I}$ 
are provided to estimate $\dot{\theta}$ and $\ddot{\theta}$ of the multibody. 

\subsection{Center of Mass Controller} \seclabel{comcontrol}

We start off with the control law proposed by Choi et al.~\cite{choi2007posture}, which realizes input-to-state 
stability by adjusting the velocity to steer the CoM $x$ back to follow the nominal velocity $\dot{x}^{ref}$ and ZMP $p_z^{ref}$ 
in the presence of errors:
\begin{equation}\eqnlabel{choicontrol} 
\begin{gathered} 
\dot{x} = \dot{x}^{ref} + \dot{x}_{ctrl},\qquad\, \dot{x}_{ctrl} = \dot{x}_{zmp} + \dot{x}_{com},\\
\dot{x}_{zmp} = K_{zmp}(p_z^{ref}-p_z),\,\dot{x}_{com} = K_{com}(x^{ref}-x).\\
\end{gathered} 
\end{equation}
We integrate its output to work with our position-based motion generator. We substitute the integral of $\dot{x}^{ref}$ with $x^{ref}$, 
while integrating $\dot{x}_{ctrl}$  with a \textit{leaky} integrator to produce the sum $x_i$.
The output position is then:
\begin{equation} \eqnlabel{integratorsum}
x^{set} = x^{ref}+x_i,\quad x_i[n] = \dot{x}_{ctrl}[n]\Delta t+(1-\alpha)x_i[n-1],\quad \alpha\in[0,1]
\end{equation}
with $\alpha$ being an exponential decay factor. This approach already provides two benefits over a non-leaky integration approach.
The first one being the ability to forget past errors, which propagate during walking from one footstep to another, while the second being
a more aggresive tuning of the $K_{zmp}$ and $K_{com}$ gains. In a non-leaky integrator, using low gain values is necessary to prevent 
velocity build-up and overshooting, at the expense of insufficient error rejection capabilities and a long settling time.
With the leaky integration however, the CoM velocity does not wind-up to levels which would first require the controller to slow it down.

This version of the controller is already capable of tracking CoM-ZMP trajectories with limited push rejection capabilities. However, 
the nature of perturbations during walking is erratic with concurrent impulse and constant forces acting on the robot, which
disrupt the rhythm of the gait. To sustain a nominal gait cycle, we introduce two additional velocity terms $\dot{x}_{eos}$ 
and $\dot{x}_{vel}$ into $\dot{x}_{ctrl}$ from \eqnref{choicontrol}:
\begin{equation}\eqnlabel{auxiliarycontrol}
\begin{gathered} 
\dot{x}_{vel} = K_{vel}(\dot{x}^{ref} - \dot{x}),\\
\dot{x}_{eos} = K_{eos}(x_{eos}^{ref} - x_{eos}).\\
\end{gathered}
\end{equation}
The reasoning behind the $\dot{x}_{vel}$ component is to maintain the reference velocity across the whole step. In standing or 
walking-on-the-spot situations this term mostly contributes to decreasing the settling time of a transient response to a push.
The second term $\dot{x}_{eos}$ uses the LIPM equations~\eqnref{lipm} with the remaining step time $t_r= (1-\mu)/f_g$ to compute reference 
and expected end-of-step~(EOS) states. Given the currently estimated state and confining the ZMP to the support polygon, the computed 
$\dot{x}_{eos}$ steadily steers the CoM towards the nominal EOS position in preparation for the next support exchange, essentially 
providing the controller with predictive capabilities. While the CoM-ZMP regulator is responsible for instantaneous adjustments 
and keeping the ZMP within stability margins as per the definition in~\cite{choi2007posture}, the introduction of extra terms 
assures consistent long-term behaviour of the system. The final form of the extended control law, which steers the CoM position to stability is:
\begin{equation} \eqnlabel{fullcontrol} 
\dot{x}_{ctrl} = \dot{x}_{com}+\dot{x}_{zmp}+\dot{x}_{vel}+\dot{x}_{eos}.
\end{equation}

\subsection{Inertia Controller} \seclabel{inertiactrl}

The reference generator and position controller are designed around the Linear Inverted Pendulum~\eqnref{lipm} concept, with dynamics
that make use of two assumptions: constant CoM height, and lack of Angular Momentum around the CoM. In practice, this leads to 
making assumptions that a fixed torso height and orientation leads to zero Angular Momentum. Building control around this discards 
a set of dynamics, which---if properly utilised---can increase the stability region and performance of the system.
There is no guarantee that the torso keeps still, especially with high joint compliance, backlash and external disturbances. 
Our inertia-based whole-body pose generation is designed to accomodate for this, along with foot swinging allowing for
unified regulation of the centroiudal system description.

The purpose of the inertia controller is to actively control the principle axes of inertia orientation $\mathbf{R_I}$ to maintain 
nominal CoM-ZMP trajectories. We do this by computing a complementary rotation $\mathbf{R_c}$ to the neutral one~\eqnref{inertiaref}:
\begin{equation} \eqnlabel{setinertia}
\mathbf{R}_\mathbf{I}^{set} = \mathbf{R}_\mathbf{c}\mathbf{R}_\mathbf{I}^{ref}.
\end{equation}
For this, we utilise the concept of \textit{augmented CoM}~(ACoM) $\tilde{x}$ introduced in~\cite{whitman2012torso}. The premise of 
the augmented CoM is to combine the linear and angular movement of the mass $x$ and flywheel $\theta_f$ of a Linear Inverted Pendulum
Plus Flywheel Model~(LIPFM) within a single state variable, and control it with respect to an Augmented ZMP $\tilde{p}_z$, putting the 
LIPFM dynamics into LIPM form:
\begin{equation} \eqnlabel{acom}
\begin{gathered} 
\tilde{x} = x + \frac{I\theta_f}{mh},\quad \tilde{p}_z = p_z + \frac{I\theta_f}{mh},\quad
\ddot{\tilde{x}} = \frac{g}{h}(\tilde{x}-\tilde{p}_z),
\end{gathered}
\end{equation}
where $m$ and $h$ are the system mass weight and height, respectively, $I$ is the inertia moment, and $\theta_f$ the flywheel angle 
around the perpendicular axis. Our aim is to maintain zero angular velocities of the flywheel, by regulating its orientation to zero. 
To achieve this, we use a similar control law as in~\eqnref{choicontrol}, and set the reference ACoM position to align with the output 
of the CoM-ZMP controller:
\begin{equation}\eqnlabel{choiangle}
\begin{gathered}
\dot{\tilde{x}}_{ctrl} = \dot{\tilde{x}}_{zmp} + \dot{\tilde{x}}_{com},\\
\dot{\tilde{x}}_{zmp} = K_{zmp}^{aug}(\tilde{p}_z^{ref}-\tilde{p}_z),\quad \dot{\tilde{x}}_{com} = K_{com}^{aug}(\tilde{x}^{ref}-\tilde{x}).\\
\end{gathered} 
\end{equation}
With the desired orientation kept at zero, the augmented and set state coincide~($\tilde{x}^{ref} \simeq x^{set}$), as well as the 
references for the regular and augmented ZMP~($\tilde{p}_z^{ref} \simeq p_z^{ref}$). Due to working in the space of the augmented 
CoM and its derivatives, the $\ddot{\theta}$ term of the Centroidal Moment Pivot~(CMP) gets hidden within $\ddot{\tilde{x}}$. 
This benefits our control scheme by enabling regulation of the CMP towards the ZMP, without having to measure or estimate 
$\ddot{\theta}$ of our abstract flywheel. The control velocity $\dot{\tilde{x}}_{ctrl}$ is then integrated as in~\eqnref{integratorsum} 
into~$\tilde{x}_i$, which holds the inertia rotation angles for $\mathbf{R_c}$:
\begin{equation} \eqnlabel{inertiaangles}
\begin{gathered}
\tilde{x}_i = \frac{I\theta_f}{mh} \Rightarrow \theta_f=\frac{\tilde{x}_imh}{I}.
\end{gathered}
\end{equation}
The rotation $\mathbf{R_c}$ can be expressed with any favorable representation. We use projected angles to avoid gimbal lock and 
maintain a set angle on two axes independently~\cite{ficht2018online}.

\subsection{Stepping Controller} \seclabel{stepping}

Conceptually, the LIPM assumes the dynamics can be split into two independent planar components. In reality, 
they are coupled by the cyclic nature of the gait. The lateral pendulum oscillations produced by instantaneous 
support foot exchanges define the rhythm of the gait. A wrongly timed foot exchange can 
quickly lead to balance loss. We introduce timing feedback, based on the nominal $y_{eos}^{ref}$ and 
expected EOS $y_{eos}$ lateral CoM error, by using the remaining step time with~\eqnref{lipm}:
\begin{equation}
\tilde{f}_g = f_g + K_t(y_{eos}^{ref}-y_{eos}).
\end{equation}
The timing adjusts the rate of step progression with $\tilde{f}_g$ in~\eqnref{gaitprogression}. Intuitively, if the 
CoM is on a trajectory to cross the lateral apex point, the progression will slow down and provide more time to both balance
controllers to bring the CoM back. Similarly, overshooting towards the swing leg will speed-up the step exchange to maintain 
the nominal cycle.

The sagittal motion is less sensitive in the timing, as the velocity is set through the step size, and known 
frequency~\eqnref{stepsize}. However, its location needs to be set accordingly, given that with the current velocity 
it is possible to return to a nominal trajectory in a finite number of steps. The active next foothold $s[k+1]$ distance $x$ is 
constantly recalculated~\eqnref{stepsize} in response to the current step EOS CoM error, smoothly adapting the foot trajectory and walking speed:
\begin{equation}
x_{s[k+1]} = v^{ref}_{g,x}/f_g - K_s(x_{eos}^{ref}-x_{eos}).
\end{equation}

\section{Experimental Results} \seclabel{results}

Verification of the proposed controller was performed on a \SI{90}{cm} tall, \iguhop robot~\cite{allgeuer2016igus} possessing 20 
position-controlled joints. The complete control pipeline was executed using the robot's on-board computer to produce 
the reference and stabilising motions. A single passthrough of the whole control loop with computing joint targets took on average 143~$\mu$s.
As the control rate is limited to \SI{100}{Hz} by the actuators, we are still left with more than \SI{98}{\%} of idle time, 
to be spent on higher-layer tasks.

For the following experiments, we set the gait frequency $f_g$ to \SI{2.6}{Hz} which is quite high, but the underpowered 
ankles necessitate making smaller step sizes. The integration decay factor $\alpha$ of all controllers was set to $0.03$, corresponding 
to a decay half time of roughly half the step time. The regular and augmented CoM-ZMP gains $K_{zmp}$, $K_{com}$, $K_{zmp}^{aug}$, $K_{com}^{aug}$ 
were set according to \cite{choi2007posture}, but with quite high gains due to the leaky integration. We have found that the long-term 
behaviour gains supplemented the controller best when set to $K_{vel} = 0.5$, $K_{eos} = 0.5K_{com}$, as the output would then equally 
mediate between the long and short-term control goals. Step timing and adjustment were set to $K_t = 10f_g$, $K_s = 1$, making a \SI{10}{cm} 
EOS error in the lateral plane fully slow down, or double the gait progression. A similar error in the sagittal direction would result in adjusting the 
foot placement by \SI{10}{cm}.

\begin{table}[!t]
\renewcommand{\arraystretch}{1.02}
\caption{Influence of the developed control techniques on walking quality. Control components are gradually added and tuned 
to achieve the best possible performance. The assessment is performed at highest achievable stable velocities with integral square errors 
of CoM, ZMP and velocity tracking, normalized by walking time and velocity.}
\tablabel{walkingquality}
\centering
\footnotesize
\begin{tabular}{l c c c c c c c }
\hline

\hline
\multicolumn{1}{c}{\bf{Control Mode}} & { } & $\quad v^{ref}_{g,x}$[$\frac{\text{m}}{\text{s}}$]$\quad$  & $\quad v_{g,x}$[$\frac{\text{m}}{\text{s}}$]$\quad$  & $\quad e_c \quad$  & $\quad e_z \quad$  & $\quad e_v \quad$   \\
\hline
\hline
\bf{Open-Loop}                      & { } & $\,$0.025$\,$  & $\,$0.1304$\,$  & $\,$5.3877$\,$  & $\,$9.3537$\,$  & $\,$76.734$\,$ & \\
\bf{Straight leg constraint}        & { } & $\,$0.050$\,$  & $\,$0.0970$\,$  & $\,$2.3155$\,$  & $\,$4.3287$\,$  & $\,$18.473$\,$ \\
\bf{Closed-Loop}                    & { } & $\,$0.005$\,$  & $\,$0.0583$\,$  & $\,$35.722$\,$  & $\,$56.002$\,$  & $\,$214.34$\,$ \\
\bf{Leaky integration}              & { } & $\,$0.125$\,$  & $\,$0.2162$\,$  & $\,$0.6266$\,$  & $\,$1.2438$\,$  & $\,$11.886$\,$ \\
\bf{Extended control terms}         & { } & $\,$0.350$\,$  & $\,$0.4403$\,$  & $\,$0.3921$\,$  & $\,$1.0304$\,$  & $\,$5.8843$\,$ \\
\bf{Inertia orientation control}    & { } & $\,$0.400$\,$  & $\,$0.4916$\,$  & $\,$0.2925$\,$  & $\,$0.8572$\,$  & $\,$3.9287$\,$ \\
\bf{Step timing regulation}         & { } & $\,$0.400$\,$  & $\,$0.5166$\,$  & $\,$0.4269$\,$  & $\,$0.9225$\,$  & $\,$4.8888$\,$ \\
\bf{Step size adjustment}           & { } & $\,$\bf{0.485}$\,$  & $\,$\bf{0.5292}$\,$  & $\,$\bf{0.1895}$\,$  & $\,$\bf{0.7522}$\,$  & $\,$\bf{1.8463}$\,$ \\
\bf{Without inertia control}        & { } & $\,$0.350$\,$  & $\,$0.3972$\,$  & $\,$0.2676$\,$  & $\,$0.7988$\,$  & $\,$3.0245$\,$ \\
\hline
\end{tabular}
\vspace{-2ex}
\end{table}

To display the influence and justify the usage of each proposed component, we perform a series of tests where a highest achievable, stable and
constant walking velocity is requested. The quality is assessed in each run by comparing the integral square errors of the relevant variable 
vector~$\mathbf{v}$ normalized by a representatively sufficient walking time $t_w$ and requested forward velocity $v^{ref}_{g,x}$:
\begin{equation}
e_{*} = \frac{1}{v^{ref}_{g,x}t_{w}}\int_0^{t_{w}}(\mathbf{v}^{ref}-\mathbf{v})^2dt,\quad * : \left.
\begin{cases}
    c\,|\,\mathbf{v}\, =\, [x\,\,y]\\
    z\,|\,\mathbf{v}\, =\, [p_{z_x}\,\,p_{z_y}]\\
    v\,|\,\mathbf{v}\, =\, [\dot{x}\,\,\dot{y}]\\
\end{cases}\right\}, 
\end{equation}

as higher velocities are much harder to maintain. The results of this assessment can be seen in~\tabref{walkingquality}. Due to the limited 
torque in the joints, coupled with backlash and compliance, the trajectories can barely be executed with frequent falls during open-loop 
walking. The straight-support-leg constraint reduced the necessary torque on the knees significantly to allow for better, more uniform 
operation. Closing the loop with a non-leaky integrator resulted in inoperability, where only a few steps caused a fall from 
self-excited instability. Introducing leaky integration provided a substantial gain in the control capabilities to the point where 
omnidirectional walking with varying velocities became possible. At moderate speeds, the walking velocity would fluctuate in
a sinusoidal manner, eventually leading to a fall. The additional control terms solved this issue, increased the accuracy and more 
than doubled the maximum speed. Reaching the limits of CoM-ZMP control, regulating the inertia orientation provided another 
substantial increase in performance. The continuous adjustments to the torso and arm positioning generated sufficient angular momentum 
to complement and relieve the CoM-ZMP controller, increased the stability margin and allowed for higher walking speeds. Finally, step size 
and timing adjustments made the gait more accomodating against sudden slips and stumbles while also compensating for visible velocity 
overshooting with a significant $e_v$ reduction. Although often omitted in typical LIPM control scenarios, here we can clearly notice
the influence that angular momentum has on walking performance. Turning off CAM regulation while keeping the step adaptation
resulted in difficulties in maintaining higher velocities, where constant gait adjustment was necessary. 

\begin{figure}[!t]
\centering{\includegraphics[width=0.85\linewidth]{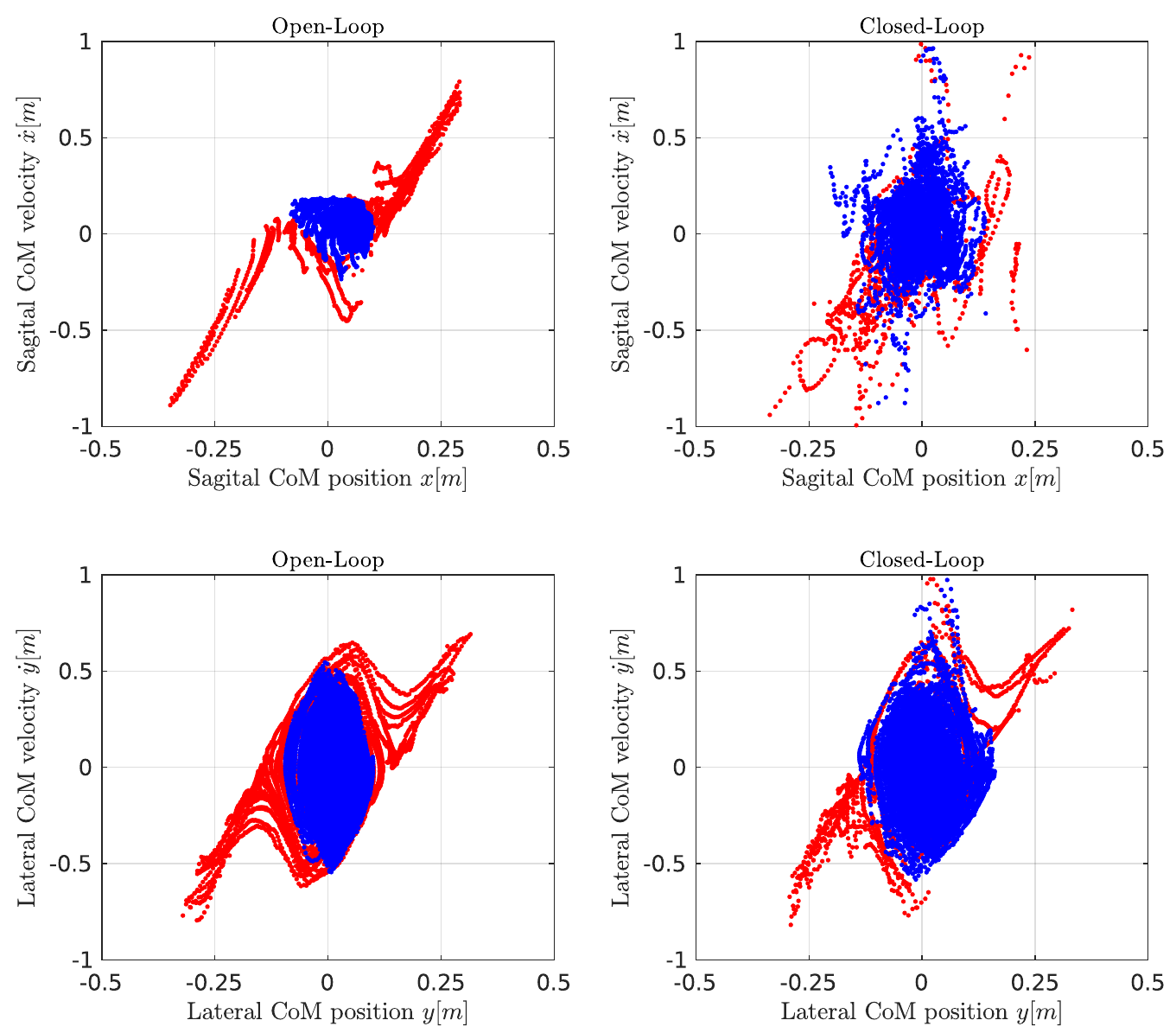}}
 \vspace{-2ex} 
\caption{Phase plots displaying the stability of open and closed-loop performance in the sagittal and lateral direction. Blue and red dots represent 
stable and unstable trajectory points respectively.}
\figlabel{results}
\vspace{-3ex}
\end{figure}

\begin{figure}[!b]
\includegraphics[width=0.495\linewidth]{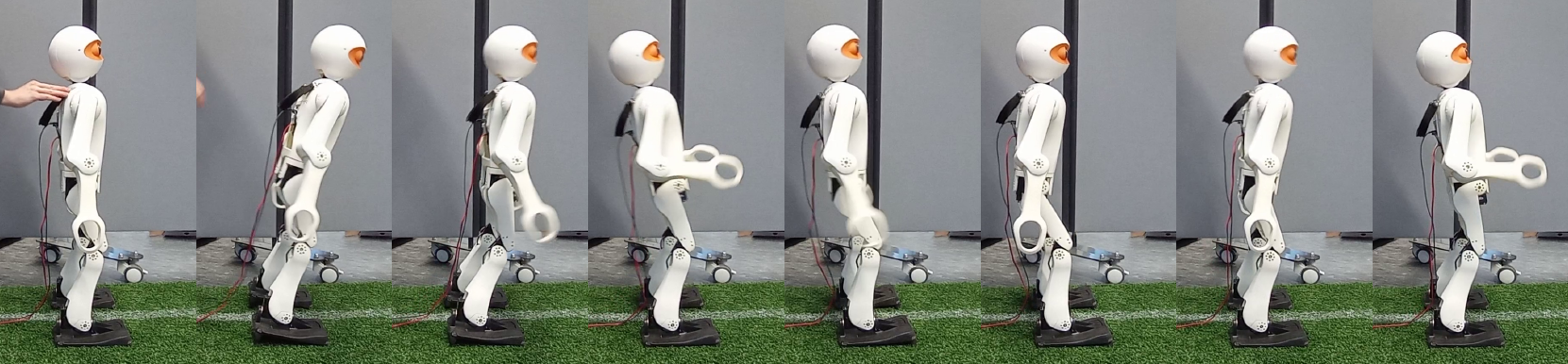}
\includegraphics[width=0.49\linewidth]{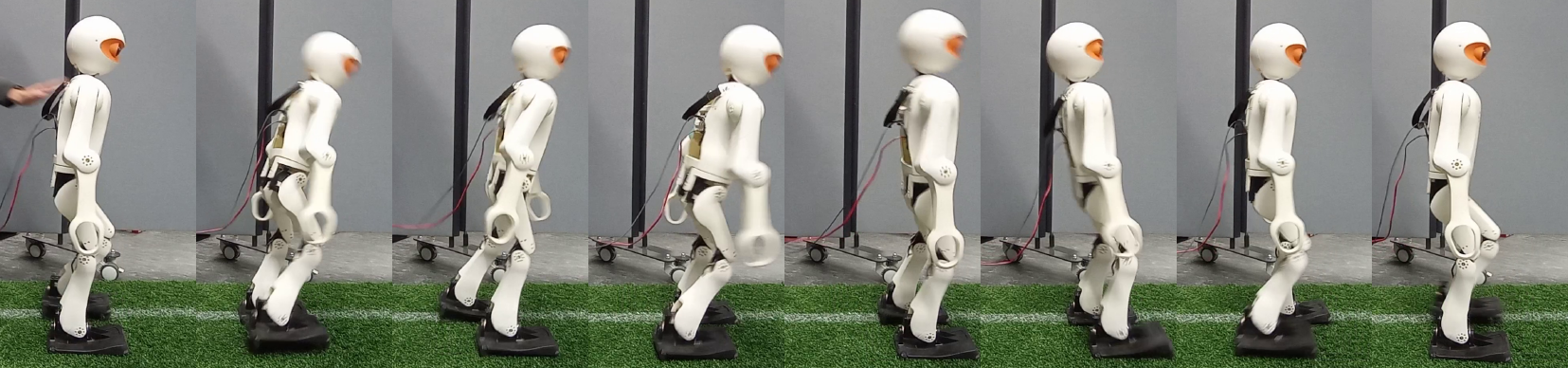}
\vspace{-2ex}
\caption{A time evolution of the robot performing push recovery with the developed controllers. 
Left: using the controller while standing. 
Right: while walking.}
\figlabel{robotdoing}
\vspace{-1ex} 
\end{figure}

We assess the push recovery performance of the presented unified control scheme by aggregating several pushes into phase plots in \figref{results}, 
with examples of robot performance shown in~\figref{robotdoing}. The Open-Loop system has a tendency to sag towards the front of the feet and innately 
dampened pushes of unsignificant strength in the sagittal direction. Closing the loop corrected the sagging and typically responded to a push with 
both leaning and stepping in it's direction. Here, an undesired coupling was observed for pushes coming from the front, where rotating the trunk would 
limit the leg touchdown towards the back due to kinematic limits. Laterally, a push on the open-loop system lead to desynchronising the gait or tipping over the outer edge of the supporting foot. Enabling feedback 
smoothly adapted the gait progression to allow the CoM and inertia controllers to bring the state back onto the nominal trajectory with a combination 
of CoM shifting and corrective inertia roll angle. A substantial stability increase in both directions is clearly visible. %

We have also evaluated the controller quantitatively in a Gazebo simulator. We compare our results to \cite{allgeuer2018bipedal},
which tackles balanced walking with a model-free, CPG-based gait. It is also the most recent work using the same robot and simulator for
the evaluation. We replicate the benchmarking scenario of performing sets of 20 pushes from a random direction with increasing
push strength. The result of this experiment is shown in \tabref{pushnumber}, with a video available\footnote{\url{https://youtu.be/MRQXE4ig0yA}}. 
Our control over the centroidal state and step feedback produces clearly better push rejection capabilities 
than~\cite{allgeuer2018bipedal}. When measuring forward velocities, a maximum mean of \SI{49.3}{cm/s} was achieved, which is also an improvement when compared to 
\SI{45.7}{cm/s} from~\cite{allgeuer2018bipedal}. Not only is the robot able to walk faster, but is twice as likely to sustain a 
relatively strong push.

\begin{table}[!t]
\renewcommand{\arraystretch}{1.02}
\caption{Number of Withstood Simulated Pushes out of 20}
\tablabel{pushnumber}
\centering
\footnotesize
\begin{tabular}{c c c c c c c c c}
\hline

\hline
\bf{Impulse} [Ns] & 1.2  & 1.5  & 1.8  & 2.0  & 2.2  & 2.4  & 2.6  & 2.8 \\
\hline
\hline
\bf{Ours} & 20  & 20  & 20  & 18  & 17  & 15  & 13  & 10 \\
\bf{Allgeuer~\cite{allgeuer2018bipedal}} & 20  & 20  & 17  & 16  & 14  & 9  & 7  & 4 \\
\hline
\end{tabular}
\vspace{-2ex}
\end{table}
\section{Conclusions}

We presented a unified control approach to whole-body locomotion and balancing of humanoid robots and 
verified it experimentally. While state-of-the-art methods using simplified models are known to work, they require sufficient sensing, 
joint torque and tracking capabilities which lower quality hardware cannot guarantee. We have shown that 
such methods are also capable of operating on such hardware, given that supplementary measures are taken.
Furthermore, we have increased the capability of an existing CoM-ZMP controller with additional feedback terms
that operate on closed-form solutions. This provides the control with predictive capabilities and greatly 
increases performance in both gait speed and stability. Unlike in MPC, our scheme easily handles
on-the-fly CoM height changes, as no preview matrices need to be computed.
The Augmented CoM concept has been employed to control the orientation of the whole-body inertia principal axes, 
which proved to be a meaningful representation for centroidal angular momentum regulation. By employing a reduced 
five-mass description of a humanoid robot, we achieved a direct feedback control scheme on the 
centroidal parameters of the system with a relatively small and intuitive set of gains. This is in contrast to current
state-of-the-art optimisation-based approaches. By adopting minimal sensing and the centroidal
system, our approach can be applied to a wide range of humanoids, even with existing hardware imperfections.

In future work, we would like to tackle issues that currently limit the hardware performance, such as 
performing the control on a latency-predicted state, anticipating kinematic limits, and modifying 
the swing foot trajectories to allow for more angular momentum compensation during a push.

\bibliographystyle{spmpsci}
\bibliography{ms}
\end{document}